\documentclass[lettersize,journal]{IEEEtran}
\usepackage{amsmath,amsfonts}
\usepackage{algorithmic}
\usepackage{algorithm}
\usepackage{array}
\usepackage[caption=false,font=normalsize,labelfont=rm,textfont=rm]{subfig}
\usepackage{textcomp}
\usepackage{stfloats}
\usepackage{url}
\usepackage{verbatim}
\usepackage{graphicx}
\usepackage{cite}
\usepackage{marvosym}
\usepackage{color}

\begin{document}

\title{Scenario Engineering for Autonomous Transportation: A New Stage in Open-Pit Mines}

\author{Siyu Teng\IEEEauthorrefmark{2}, Xuan Li\IEEEauthorrefmark{2}, Yuchen Li, Lingxi Li, ~\IEEEmembership{Senior Member,~IEEE,} Zhe Xuanyuan\textsuperscript{\Letter},\\ Yunfeng Ai\textsuperscript{\Letter},  Long Chen\textsuperscript{\Letter}, ~\IEEEmembership{Senior Member,~IEEE,}
        % <-this % stops a space
\thanks{Our work was supported in part by the National Natural Science Foundation of China under Grant 62203250, in part by the Young Elite Scientists Sponsorship Program of China Association of Science and Technology under Grant YESS20210289, in part by the China Postdoctoral Science Foundation under Grant 2020TQ1057 and Grant 2020M682823, in part by the Guangdong Provincial Key Laboratory of Interdisciplinary Research and Application for Data Science, BNU-HKBU United International College 2022B1212010006, in part by Guangdong Higher Education Upgrading Plan with UIC research grant R0400001-22 and R201902. (Siyu Teng and Xuan Li contributed equally to this work). (Corresponding authors: Zhe Xuanyuan, Yunfeng Ai, and Long Chen).

Siyu Teng and Yuchen Li are with Hong Kong Baptist University, Kowloon, Hong Kong, 999077, China and BNU-HKBU UIC, Zhuhai, China (e-mail: siyuteng@ieee.org).

Xuan Li is with the Department of Mathematics and Theories, Peng Cheng Laboratory, Shenzhen 518000, China (e-mail: lix05@pcl.ac.cn).

Lingxi Li is with the Purdue School of Engineering and Technology, Indiana University-Purdue University Indianapolis (IUPUI), Indianapolis, USA. (e-mail: ll7@iupui.edu)

Zhe Xuanyuan is with Guangdong Provincial Key Laboratory IRADS, BNU-HKBU UIC, Zhuhai, 519087, China (e-mail: zhexuanyuan@uic.edu.cn).

Yunfeng Ai is with the University of Chinese Academy of Sciences, Beijing, 100049, China (e-mail: aiyunfeng@ucas.ac.cn).

Long Chen is with the State Key Laboratory of Multimodal Artificial Intelligence Systems and the State Key Laboratory of Management and Control for Complex Systems, Chinese Academy of Sciences, Beijing, 100190, China. Long Chen is also with WAYTOUS Inc., Beijing, 100083, China, and the Guangdong Laboratory of Artificial Intelligence and Digital Economy (SZ), Shenzhen 518107, China (e-mail: long.chen@ia.ac.cn).
}% <-this % stops a space
\thanks{Manuscript received April 19, 2021; revised August 16, 2021.}}

% The paper headers
\markboth{Journal of \LaTeX\ Class Files,~Vol.~14, No.~8, August~2021}%
{Shell \MakeLowercase{\textit{et al.}}: A Sample Article Using IEEEtran.cls for IEEE Journals}

% \IEEEpubid{0000--0000/00\$00.00~\copyright~2021 IEEE}
% Remember, if you use this you must call \IEEEpubidadjcol in the second
% column for its text to clear the IEEEpubid mark.

\maketitle

\begin{abstract}
In recent years, open-pit mining has seen significant advancement, the cooperative operation of various specialized machinery substantially enhancing the efficiency of mineral extraction. However, the harsh environment and complex conditions in open-pit mines present substantial challenges for the implementation of autonomous transportation systems. This research introduces a novel paradigm that integrates Scenario Engineering (SE) with autonomous transportation systems to significantly improve the trustworthiness, robustness, and efficiency in open-pit mines by incorporating the four key components of SE, including Scenario Feature Extractor, Intelligence and Index, Calibration and Certification, and Verification and Validation. This paradigm has been validated in two famous open-pit mines, the experiment results demonstrate marked improvements in robustness, trustworthiness, and efficiency. 
\textcolor{black}{By enhancing the capacity, scalability, and diversity of autonomous transportation, this paradigm fosters the integration of SE and parallel driving and finally propels the achievement of the `6S' objectives.}
\end{abstract}


\section{Introduction}
\IEEEPARstart{O}{pen-pit} mining is manifested as a primary method for mineral extraction, focusing mainly on near-surface mineral. Compared to the prevalent underground mining, this approach reduces the costly infrastructure in underground operations, such as coal cutters, hydraulic props, and various conveyors. Owing to its lower costs and enhanced safety protocols, open-pit mining plays a crucial role in meeting global mineral requirements \cite{huang}. The broad operational space of open-pit mines allows for the implementation of various specialized, large-scale machinery such as mining trucks, backhoe loaders, and electric shovels, enhancing overall productivity. These extensive ranges of controllable machinery pave the way for future autonomous transportation in open-pit mines.

\begin{figure}[t]
\centerline{\includegraphics[width=0.65\linewidth]{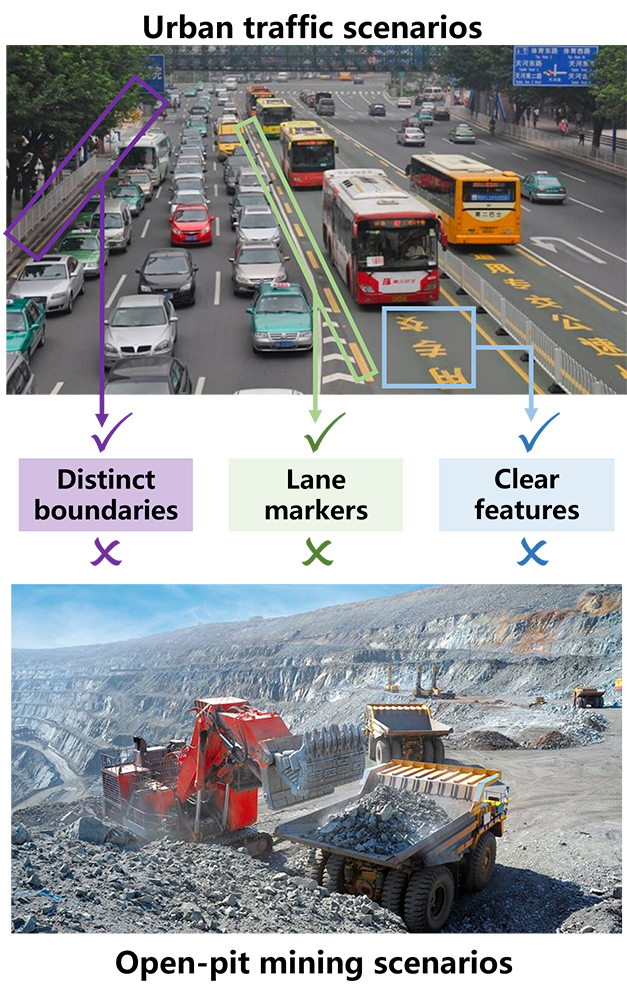}}
\caption{Typical differences between open-pit mining scenarios and urban traffic scenarios.}
\label{openpit_mine}
\end{figure}

Mineral transportation constitutes a significant portion of the total production process in open-pit mines, enhancing the efficiency of mining trucks is the most important for bolstering the productivity of open-pit mines \cite{libai2}. To this end, a multitude of researchers dedicated their research to the implementation of autonomous transportation in open-pit mines \cite{libai1,gao}. \textcolor{black}{However, the harsh environment and complex conditions complicate the implementation of automated systems. Most open-pit mines are located in remote areas and always suffer from climatic adversities such as blizzards, sandstorms, and heavy rainfalls.} These conditions severely distort the trustworthiness and robustness of the automated systems, leading to potential danger from false positive detection. \textcolor{black}{Moreover, as shown in Fig. \ref{openpit_mine}, compared with the urban scenario, the mining scenario often presents challenges with its homogenized scenario, such as vague boundaries, Fuzzy features, and no lane markers \cite{teng}.} This homogeneity can hinder the feature extraction process, making convolutional networks less effective in capturing latent features. \textcolor{black}{Such limitations may lead to unpredicted decisions by autonomous vehicles, especially some events outside of the training dataset, which are referred to as Out-of-Distribution (OOD) events for convenience.} Additionally, mining trucks frequently interact with other heavy machinery like electric shovels and excavators. \textcolor{black}{The margin of collision often within mere centimeters in these interactions, further emphasizes the complexity of ensuring accurate vehicular responses}. Therefore, formulating and refining the trustworthiness and robustness of autonomous transportation systems becomes a hard task in open-pit mines \cite{HE2}.

To overcome the challenges in open-pit mines, this research introduces an innovative autonomous transportation paradigm, which is grown from our previous research \cite{dtpi}. This paradigm enhances the trustworthiness, robustness, and efficiency of the autonomous transportation system in open-pit mines by infusing scenario feature extractor, Intelligence and Index (I\&I), Calibration and Certification (C\&C), and Verification and Validation (V\&V) within the SE framework \cite{selfcitation}.

\section{relate works}

\subsection{Open-pit mines}
\textcolor{black}{As illustrated in Fig. \ref{openpit_mine}, open-pit mining manifests as an extraction technique mainly implemented for near-surface minerals.} Since the beginning of the twenty-first century, the extraction technologies of open-pit mining have significantly matured, particularly in mechanized extraction. This advancement equips each production process with specialized machinery, thereby reducing the inherent risks of open-pit mines.

Open-pit mining operations are marked by their transparent procedures, aiding real-time management to ensure safety and efficiency in mineral extraction \cite{yang2}. This extraction method boasts exceptional adaptability, allowing for timely adjustments of mining strategies in response to potential shifts in mineralogical geology, ensuring an optimal harnessing of mine resources \cite{mining5}. Despite these advances, open-pit mining still suffers from high work risks, low levels of intelligence, and harsh working conditions. Efficient extraction in open-pit mines necessitates the cooperative operation of various machinery, notable for their excellent loading capacity and terrain adaptability, mining trucks have emerged as the dominant transportation equipment within open-pit mines \cite{libai1}. Incorporating autonomous transportation systems into these trucks could significantly boost productivity and safety, the advantages of autonomous mining trucks are listed below:

\begin{itemize}
    \item Autonomous mining trucks significantly enhance operational safety and efficiency for mineral transportation. The collaborative operation of multiple large-scale specialized machinery introduces potential hazards for engineers. By deploying autonomous transportation methods, potential risks stemming from humans can be efficaciously reduced, ensuring secure interactions with other machinery during transportation. 
    
    \item Autonomous mining trucks integrated with non-visual perception technologies, such as LIDAR, Radar, and infrared cameras, can facilitate round-the-clock operations by executing intricate control commands, guaranteeing continuous production in open-pit mines, thus significantly enhancing transportation efficiency. 
\end{itemize}

\textcolor{black}{Compared to structured scenarios, such as urban traffic and logistics parks, the unique properties of open-pit mines present a series of challenges for autonomous transportation \cite{DRT}.} Structured scenarios offer plenty of traffic elements and scenario features. However, the retaining walls and drivable areas in open-pit mines are composed of homogeneous materials, lacking distinct boundary characteristics. Furthermore, urban scenarios have a wide variety of transportation participants, whereas open-pit mining has a more limited variety, which only offers sparse features due to the homogeneity of the scenarios. To address the challenges in boundary blurring and features lacking, this research proposes an autonomous transportation paradigm based on SE, ensuring a more trustworthy and robust automatic system.

\subsection{Autonomous Driving Framework}
\subsubsection{Pipeline Framework}
The pipeline framework decomposes the comprehensive autonomous driving architecture into well-defined sub-tasks, such as perception \cite{automine,52_2,52_1}, prediction \cite{huang111}, decision-making, planning \cite{libai3,tmatch}, and control \cite{zhang2023vehicle}, with each sub-tasks operating in relative isolation. The obvious advantages of this framework are clear intermediate representations and deterministic decision processes, which facilitate locating the root of error when an autonomous vehicle is stuck into OOD events, thereby enabling in-depth reasoning of abnormal specific control commands. However, the pipeline framework has its intrinsic challenges \cite{motionplanning}. Firstly, a well-designated sub-task only searching for its local optimal policy, might not be suitable for the total pipeline framework. The superiority at the sub-task level does not guarantee global optimality, thus limiting the generalizability of the pipeline framework. Additionally, the decoupled chaining of multiple sub-tasks implies that a malfunction within a singular task can impact the whole framework, the robustness of which still needs to be improved. Lastly, these sub-tasks are always coupled with several human-craft heuristic constraints,  the frequent interactions might promise the responsiveness of the pipeline framework \cite{teng}.

\subsubsection{End-to-End Framework} \label{e2eplanning}
The end-to-end framework leverages a data-driven method, interpreting the total driving task from raw perception data to control signal as an integrated optimal mapping problem \cite{motionplanning,56}. Through the excellent fitting ability of neural networks, this framework identifies and learns the latent intricate relationship. Such a novel framework avoids potential feature attenuation and inductive biases by human-defined intermediate constraints, thereby enhancing its adaptability and generalizability across diverse driving scenarios. Typically, the end-to-end framework is always designed by singular or multiple neural networks, showcasing advantages in robustness and response speed compared to the pipeline framework. However, an intrinsic limitation of this framework is the uninterpretability property. The absence of clear intermediate explanations makes it hard to locate errors and understand specific driving policies. 

\textcolor{black}{To ameliorate the uninterpretability issue, many researchers \cite{oriplanning} are exploring the modular end-to-end planning framework, which breaks down the total driving task into interconnected modules, emphasizing downstream tasks to enhance interpretability \cite{Sun, hierar}.} In the development of differentiable perception modules for modular end-to-end autonomous driving, the selection of appropriate loss functions and the screening of suitable modules for different scenarios are critical issues in the implementation of the method \cite{tip}.

\section{Scenario Engineering for Autonomous Transportation}

Recent trends in Artificial Intelligence (AI) indicate an escalating interest in SE \cite{SELetter} and its emerging applications \cite{SE1,SE2}, as evidenced by a growing number of studies focusing on their applications and implications. It focuses on creating, analyzing, and supervising different scenarios of artificial systems by data-driven methods. By reasoning potential relationships from the holistic perspective in certain scenarios, SE aims to enhance the trustworthiness and robustness of AI systems \cite{10264145}. This comprehensive method deeply supervises perception data and evaluates well-trained models to identify potential risks in the autonomous driving framework, improve training efficiency, and validate the trustworthiness and robustness of well-trained models. 

%. In addition, feature engineering often proves insufficient to handle the complexity and diversity of scenarios, and its black-box effect leads to an inability to accurately pinpoint the stem of errors and target improvements when accidents occur.

\begin{figure*}
    \centering
    \includegraphics[width=0.85\linewidth]{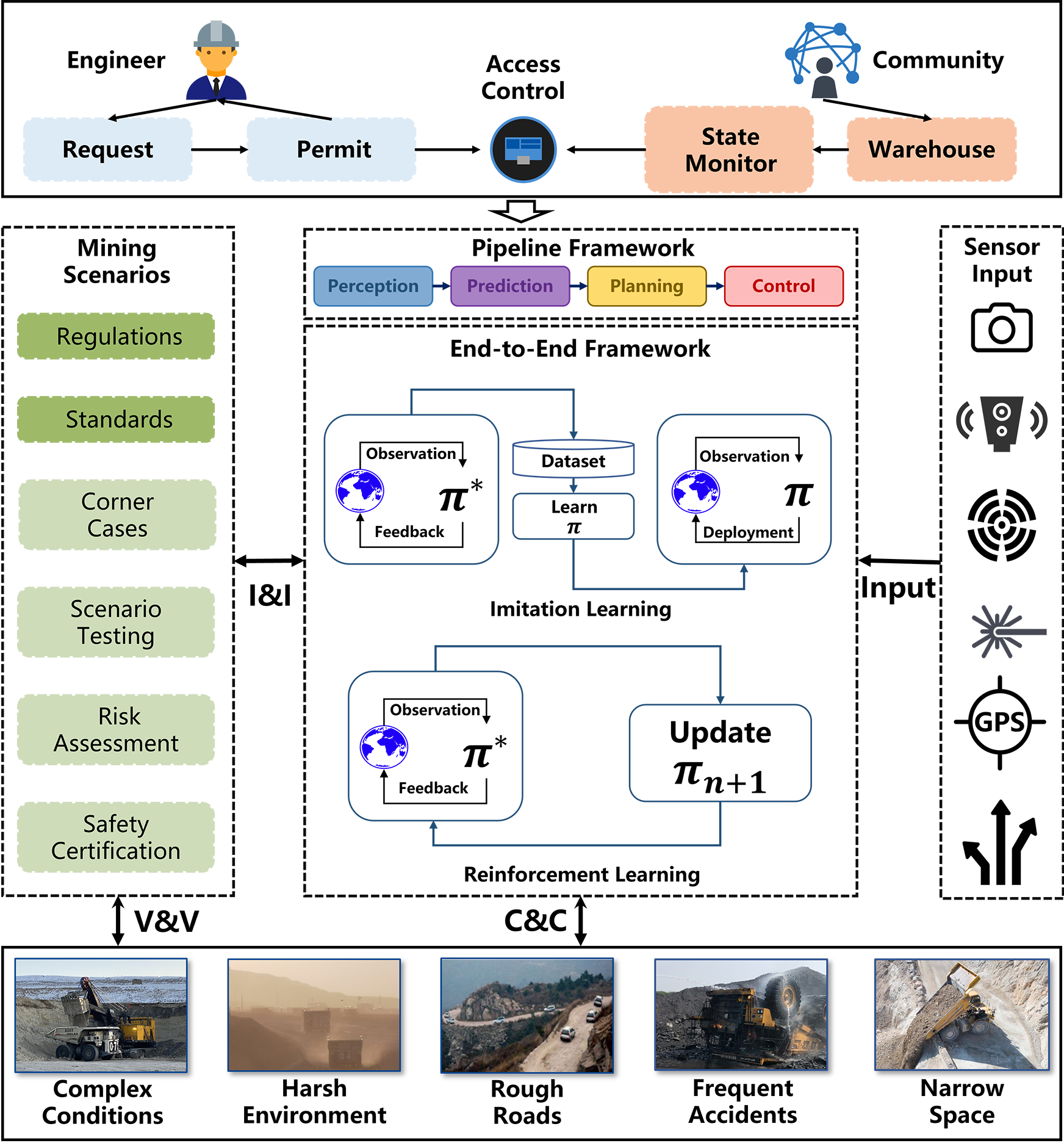}
    \caption{The framework of autonomous transportation for open-pit mines, which integrated Scenario Engineering to improve the robustness and trustworthiness of the transportation model. SFE processes all potential input data from the whole scenario perspective, \textcolor{black}{ranging from most multi-source sensor fusion methods used in the perception phase to complex high-level commands employed in the planning phase. I\&I is proposed for data argumentation and validation in autonomous transportation in open-pit mines. C\&C provides a novel methodology to ensure efficient learning efficiency of the unmanned transportation model as well as the effective criterion of the model’s performance. V\&V involves ensuring the correctness and performance of autonomous transportation models at open-pit mines.}}
    \label{fig:framework}
\end{figure*}

\textcolor{black}{Traditionally, the development of AI heavily relies on feature engineering, which builds a solid cornerstone for machine learning \cite{resnet}. However, this method is always criticized because of neglected features and thus results in low generalization resulting in complex scenarios.} By considering the interactions of agents from the perspective of the entire scenario, and supplemented with intelligent configuration methods and efficient validation approaches, SE greatly enhances the trustworthiness and robustness of AI systems crossing various real and virtual scenarios. Especially in safety-focused open-pit mines, these qualities are critically important. As shown in Fig. \ref{fig:framework}, integrating SE into autonomous transportation for open-pit mining involves four parts, all designed to enhance the robustness, trustworthiness, and learning capacity of the trained model. These parts include Scenario Feature Extractor (SFE), Intelligence \& Index (I\&I), Calibration \& Certification (C\&C), and Verification \& Validation (V\&V).

\subsection{Scenario Feature Extractor}

Feature engineering plays a pivotal role in machine learning, which is evident in feature extraction at various layers of the pipeline framework and relationship mapping within the end-to-end framework \cite{resnet, Liu1}. Although much research \cite{wang1,automine} focuses on finding certain effective methods to discern the potential relationships within raw perception data, an important underexplored problem is how to create adaptable and generalizable constraints from the heavy reliance on explicit features. Furthermore, the dependence on human-defined labels introduces subjective biases, which can propagate between the latent layers of networks, potentially resulting in faulty decisions. Lastly, with increasing data volume and problem complexity, feature engineering presents scalability issues, struggling to promptly capture and update features in evolving scenarios \cite{nips}. This is particularly problematic in unstructured environments like open-pit mines, making it difficult to capture all potential relationships and update features promptly to adapt to changing scenarios. These limitations might potentially lead to serious accidents for intelligent vehicles, especially in unstructured scenarios such as open-pit mines with fuzzy boundaries and sparse features. 

To ameliorate these issues, SE provides a comprehensive method, named scenario feature extractor (SFE), to precisely identify the interactions and cooperation between specialized machinery, even in the complex and hostile scenario of open-pit mines, as shown in Fig. \ref{fig:framework}. Unlike traditional feature engineering, SE emphasizes the comprehensive understanding of the overall scenario features, enhancing adaptability, generality, and fairness. The overarching object of SE is to mitigate the potential bias, augment system transparency, and enable efficient scalability through automated knowledge technology.

SE enhances data-driven models with comprehensive learning methods including optimizing the volume, representation, and distribution of datasets, enabling these models to effectively learn and address complex challenges. SE extracts features from the whole scenario perspective, enhancing AI models with adaptability and generalizability, especially in novel situations. Another benefit of SE is minimized bias of datasets, promoting fairness, transparency, and accountability. In addition, SE integrates automated knowledge to alleviate manual feature selection, allowing the system to adapt better to challenging scenarios \cite{knowledgeautomation}.

%One of the significant advantages of SE is minimized bias during the training process. By including a broader array of contexts and factors, SE promotes fairness, transparency, and accountability. Furthermore, SE integrates automated knowledge, which alleviates the burden of manual feature selection, allowing the system to adapt to challenging situations. This alleviation the limitations of human influence in data-driven models. In contrast with traditional feature engineering, which may encounter limitations in new scenarios, SE methods demonstrate better adaptability, ensuring increased levels of fairness, transparency, and interpretability.

%In contrast to conventional feature engineering, which may encounter limitations in unfamiliar scenarios, SE methods showcase superior adaptability. This adaptability guarantees higher levels of fairness, transparency, and interpretability, making SE an invaluable tool in the realm of data-driven methods.

The introduction of Transformer \cite{transformer} offers a practical approach for implementing SFE. Utilizing self-attention and inter-attention mechanisms, it identifies associative relationships within input sequences without being constrained by a fixed perceptual distance \cite{ViT,swintransformer, Transfuser2}. Compared with traditional neural networks, Transformer captures extensive dependencies in raw data with heterogeneous structures, providing a better understanding method for the surrounding environment. When processing sequence data, Transformer evaluates the entire sequence simultaneously, further enhancing its ability to grasp intricate patterns and relationships \cite{pointformer}. In addition, Traditional neural networks augment the perceptual field by overlaying numerous latent layers, which may lead to feature loss or blurring, and result in poor performance in scenarios with sparse features. \textcolor{black}{To ameliorate this problem, several Transformer-based methods have been incorporated in SFE to encode input data from various perceptual data sources in both pipeline framework and end-to-end framework.}

Both two frameworks access multi-model inputs such as raw perception data, vehicle states, and high-level commands. SFE covers all potential input types for both frameworks, from multi-source sensor fusion methods used in the perception layer to complex high-level commands employed in the planning layer. Considering that the format of the raw sensory data may be heterogeneous, the SFE integrates multiple transformers to process and manage multi-modal information.

In pipeline framework, for mostly visual inputs, Transformer-based models such as ViT \cite{ViT}, Swin Transformer \cite{swintransformer}, PoolFormer \cite{pointformer}, and some lightweight variants like EfficientFormer \cite{Efficientformer}, MobileFormer \cite{Mobileformer}, and EdgeViTs \cite{EdgeVit} are embedded within SFE. They are designed to produce potential representations from multiple perspectives. In cases of point cloud data, PCT \cite{Pct} and PointFormer \cite{pointformer} are utilized to create point or voxel embeddings. For straightforward scalar data, which includes the states of autonomous vehicles and high-level navigation commands, the fundamental Transformer model \cite{transformer} suffices for the learning process. SFE employs selective activation of specific Transformer models based on the type of raw perception data input, such as vision-only or multi-source data. For vision-only inputs \cite{ViT}, the backbone network of visual transformers is utilized. For multi-source perceptual input, modality-specific transformers are combined, which is followed by a designed fusion layer, that supports multiple fusion methods: point-level, pixel-level, and BEV-level fusion. Point-level fusion maps image features onto 3D points or voxels using 3D-2D matching. Pixel-level fusion combines depth or distance features from point cloud data with pixel features. Meanwhile, BEV-level fusion establishes a universal bird's eye view space, enabling the consistent representation and fusion of images and point clouds.

\begin{figure*}[t]
    \centering
    \includegraphics[width=0.85\linewidth]{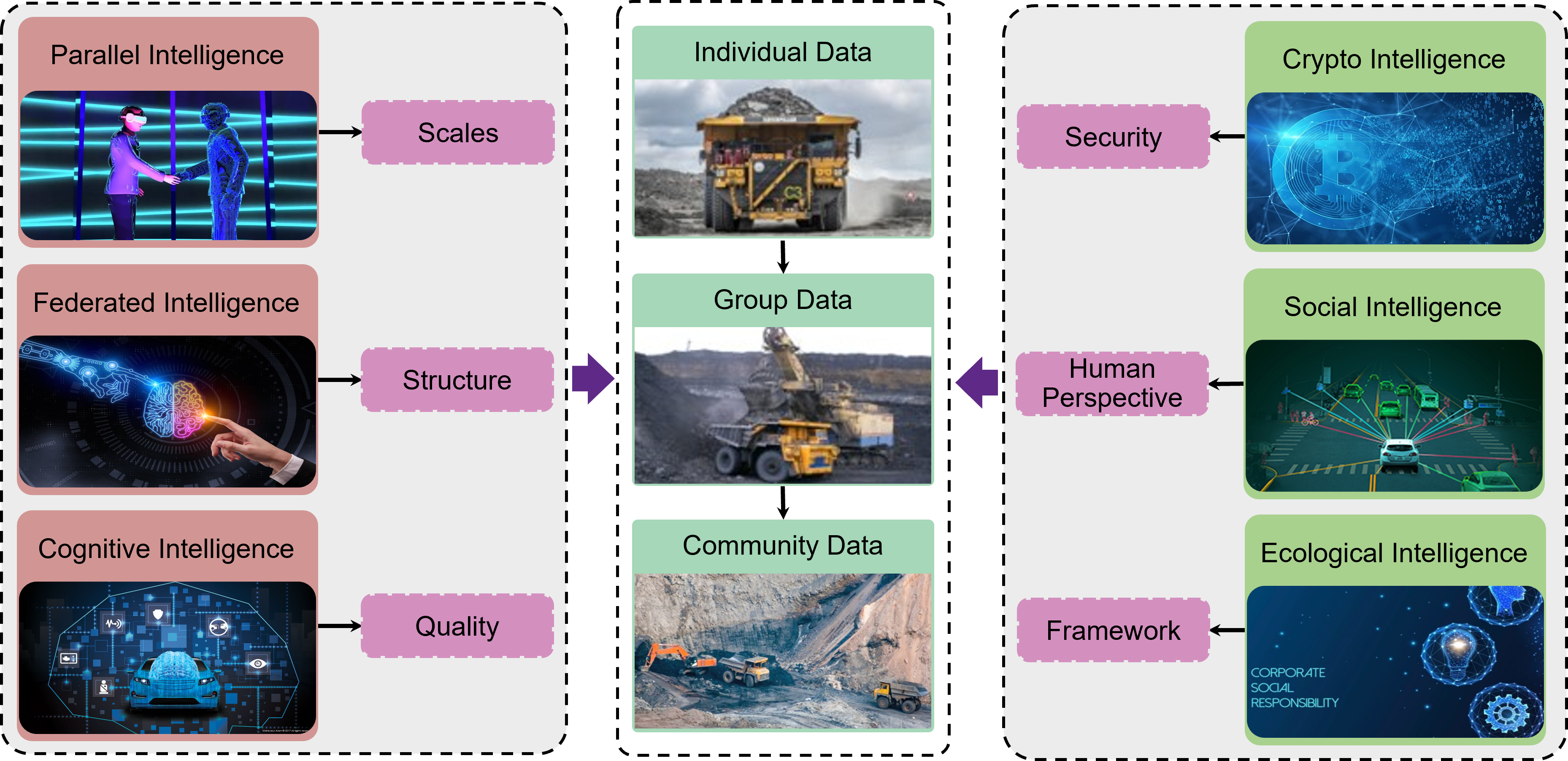}
    \caption{Intelligence \& Index for data argumentation and validation in open-pit mines.}
    \label{fig:IandI}
\end{figure*}

In the end-to-end framework, SFE combines the scenario feature with the capabilities of attention models in both imitation learning and reinforcement learning.

In imitation learning, due to the variety of sensor configurations available within a reasonable computational budget, researchers have focused their attention on employing more modalities and advanced architectures to capture correlative and representative features in surrounding environments, such as TransFuser \cite{Transfuser} and many variants \cite{Transfuser1}. However, as stated in Section \ref{e2eplanning}, the end-to-end approach has limitations in terms of interpretability due to the black-box property. To address this issue, SEF integrates plenty of advanced methods such as NEAT \cite{neat} and HIIL \cite{hierar}. These methods employ auxiliary modules to supervise the learning process to improve interpretability. In Reinforcement Learning (RL), online and offline training are two main training modalities with learning-by-trial-and-error. Some researchers \cite{TIVLeader} show an efficient online training model for SEF by lightweight an attention-based model. Decision Transformer \cite{decisiontrans} and Trajectory Transformer \cite{trajectorytrans} are two formulate autonomous driving tasks as a single sequence modeling problem for offline learning. Decision Transformer \cite{decisiontrans} is a model-free method conditional on rewards, which learns a Transformer-based policy that takes into account historical states and actions as well as goal reward and outputs of the selected actions. Trajectory Transformer \cite{trajectorytrans} is a model-based method that trains a planning model based on a mutual attention mechanism that illuminates potential trajectories in scenarios. 

To conclude, for the effective development and deployment of autonomous mining trucks in open-pit mines, the adoption of SFE is essential, which aids in conveying complex and evolving scenario information. The comprehensive, adaptable nature of SFE, along with its automation capabilities, fosters transparency, fairness, and accountability. This guarantees efficient and ethically compliant autonomous transportation in the dynamic environment of open-pit mines.
% -------------------------
\subsection{Intelligence \& Index}

With the development of AI, the importance of data quality and security has increasingly come to the forefront, especially in high-risk environments such as open-pit mines \cite{17}. Concerning data quality, existing datasets hardly encompass all transportation scenarios, which severely limits the robustness and trustworthiness of autonomous transportation systems \cite{yutong12}. Moreover, due to the uneven distribution of data, rare but high-risk accidents prove impossible for data-based models to effectively learn and predict reasonable outcomes. Ensuring the robustness and trustworthiness of autonomous mining trucks in "Corner Cases" poses an urgent challenge.

\begin{figure}[t]
    \centering
	  \subfloat[Turn right.]{
       \includegraphics[width=0.48\linewidth]{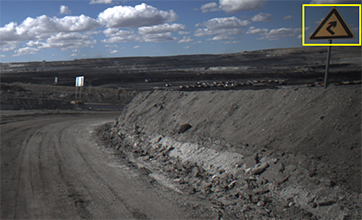}}
    \label{1a}\hfill
	  \subfloat[Speed limit 50km/h]{
        \includegraphics[width=0.48\linewidth]{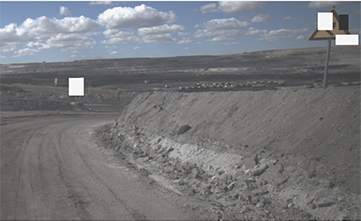}}
    \label{1b}\hfill

	  \subfloat[Pedestrian detected]{
        \includegraphics[width=0.48\linewidth]{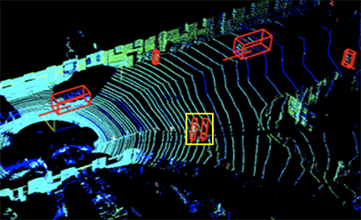}}
    \label{1c}\hfill
	  \subfloat[Pedestrian ignored]{
        \includegraphics[width=0.48\linewidth]{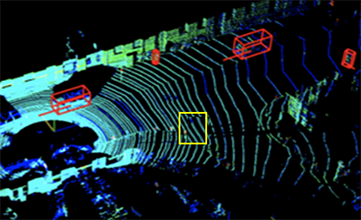}}
     \label{1d} \hfill
	  \caption{Both visual sensors (e.g. cameras) and non-visual sensors (e.g. LIDAR) are susceptible to processing perturbations. Even the slightest ones can yield an unimaginable influence on computational results, thereby leading to potentially catastrophic incidents in autonomous transportation in open-pit mines.}
	  \label{fig:affect} 
\end{figure}

\textcolor{black}{In terms of data security, autonomous vehicles are facing an increasing threat from hacking attempts. As shown in Fig. \ref{fig:affect}, even the slightest perturbations in perception data could result in profound output biases. To enhance the trustworthiness and robustness of autonomous transportation in open-pit mines, I\&I is dedicating efforts to improve the quality of training data through volume, distribution, and security.}

\textcolor{black}{I\&I includes `6 Intelligence'  and `6 Index'. As shown in Fig. \ref{fig:IandI}, the `6 Intelligence' serves as the fundamental element of SE, whose primary purpose is to augment and validate the dataset.} These intelligence consist of Cognitive Intelligence \cite{knowledgeautomation}, Parallel Intelligence \cite{parallelplanning}, Encryption Intelligence, Joint Intelligence, Social Intelligence \cite{18}, and Ecological Intelligence\cite{yutong14}. To complement the specific functions of the goal of SE, corresponding indices are meticulously crafted, which serve the purpose of evaluating and expressing the effectiveness of SE and together form the `6 Index'. These indices include the Safety Index, Security Index, Sustainability Index, Sensitivity Index, Service Index, and Smartness Index. Each index is designed to evaluate a specific content of SE, ensuring a comprehensive and nuanced assessment of its impact and functionality.

To ensure the quality of the dataset for model training in autonomous transportation models, `6 Index' enhances data quality from multiple dimensions. As depicted in Fig. \ref{fig:IandI}, the `3 Intelligence', including Cognitive Intelligence, Parallel Intelligence \cite{yutong1}, and Federated Intelligence, focus on improving the volume and distribution of the dataset, which significantly enhances the quality of training data, effectively alleviating issues related to volume and uneven distribution. The remaining `3I', including Encryption Intelligence, Cognitive Intelligence, and Ecological Intelligence, are utilized for data security validation, whose primary function is to guarantee the trustworthiness of multi-level data (Individual Data, Group Data, and Community Data) in open-pit mining scenarios, reduce data biases, and further enhance the overall quality of the dataset.

%%%%%%%%%%%%%%%%%%%%%%%%%%%%%%%%  ---------------------------
\subsubsection{`3I' for data augmentation}
\
\newline \indent \textit{Parallel intelligence}: It ensures data quality from a distribution perspective, by converting small independent data into large coupled data \cite{18}. It can facilitate data augmentation by modeling artificial systems and executing computational experiments \cite{19}. The process of parallel intelligence is as follows: first, a high-precision and fully dynamic artificial system is constructed corresponding to a real physical system; next, autonomous transportation data is trained, predicted, and evaluated through computational experiments and parallel execution by parallel system theory; finally, these data is controlled and managed by establishing interactive and mutually beneficial relationships between the real physical system and the virtual artificial system. With the continuous interaction between virtual and reality, parallel intelligence can push the artificial system closer to the real system so that the quality of the generated data will be higher \cite{10242366}. The improved dataset derived from this process can be utilized to train and validate the AI model to improve trustworthiness and generalization.

\begin{figure}[b]
    \centering
    \includegraphics[width=0.85\linewidth]{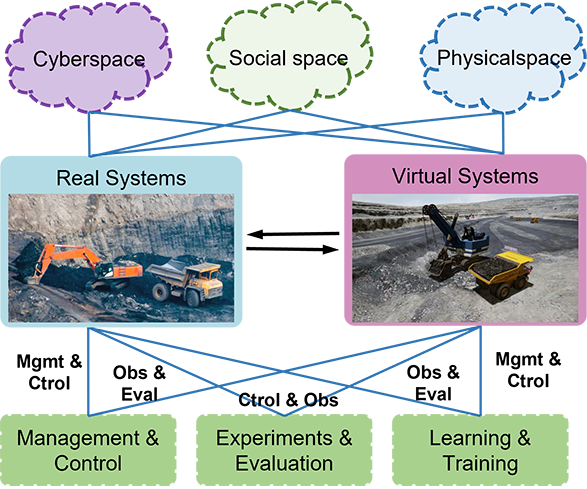}
    \caption{ACP-based parallel intelligence framework for autonomous transportation in open-pit mines.}
    \label{fig:parallelintelligence}
\end{figure}

\textit{Cryptographic intelligence}: It enables data validation from a security perspective. Most data-driven methods are vulnerable to advanced hacking techniques \cite{SE1}, especially in autonomous transportation at open-pit mines with high-demand for safety levels. To address the growing challenge of cybersecurity, extensive research \cite{ huang1} is focused on adversarial crypto intelligence. The objective is to develop resilient machine learning models that can withstand diverse adversarial scenarios and establish trustworthy defenses for cyber communications \cite{Yuke1}. Several mathematical AI models \cite{AdaTest}, including variants of reinforcement learning and imitation learning, play a crucial role in this area. These models demonstrate how attack vectors exploit vulnerabilities in data-driven models and provide avenues for several cyber defense policies to safeguard these models from various attack techniques. Moreover, in the realm of data validation, crypto intelligence emerges as a pivotal tool to ensure the integrity and authenticity of training datasets through a secure and transparent framework. Blockchain technology \cite{lijuanjuan1, lijuanjuan2}, with its decentralized nature and efficient encryption methods, provides robust support for the implementation of crypto intelligence. Its properties not only ensure data security and transparency but also enhance data immutability and verifiability. This enhances the robustness and trustworthiness of cryptographic intelligence, thereby bolstering the overall security posture in the data validation process.
%1024

\textit{Federated Intelligence}: It achieves data enhancement from a structural perspective. As shown in Fig. \ref{fig:IandI}, Federated Intelligence transforms individual data of a single organization into joint data of multiple groups. It divides all the data into three levels: individual-level (data of a single special machine such as mining truck, excavator, etc.), group-level (data of collaborative operation such as excavator-truck, etc.), and community-level (production data of the transportation road, loading/unloading area). The multi-level data architecture utilizes small data to generate big data, and then refines them to form deep intelligence, which effectively breaks through the limitations of traditional methods and solves the tension between complex phenomena and convergent solutions in the transportation scenario of open-pit mines. Federated intelligence facilitates data federation through federated control and service via federated management \cite{Yuke2}. This operation is governed by federated contracts, consensus, incentives, and security, providing a robust framework for seamless data integration and analysis.

\subsection{“3I” for data validation}
\
\newline \indent \textit{Cognitive Intelligence}: \textcolor{black}{It enhances data from a quality perspective, with a focus on improving the matching level of data-driven methods to the current dataset. Cognitive Intelligence plays a crucial role in data augmentation by integrating human factors in the training and testing stages of autonomous transportation systems \cite{parallelintelligence}. It forms a hybrid intelligence paradigm with human-in-the-loop where humans continuously interact with intelligent systems \cite{zhuzhengqiu1}.} When the output of data-driven models shows low confidence, human intervention is employed to adjust parameters and provide reasonable solutions, which interactive process creates a feedback loop, enhancing the intelligence level of the autonomous system. Furthermore, Cognitive Intelligence merges human cognition and enables humans to analyze and address complex issues, fostering mutual adaptation and collaboration. This symbiotic relationship leads to bi-directional information exchange and control. By combining human cognition with computer capabilities, Cognitive Intelligence optimizes data-driven methods, enhancing data preprocessing and learning quality.

\textit{Social Intelligence}: It validates data from the human-recognized perspective of social signals and social relationships, addressing complex decision-making issues by focusing on social behavior and information propagation \cite{yutong13}. Social Intelligence leverages the nearly unlimited data and information in the cyber world, overcoming resource constraints and temporal limitations of the physical world. In the context of data validation of autonomous transportation systems, Social Intelligence is manifested as a critical approach in analyzing the social developments related to the data generation and validation process. As shown in Fig. \ref{fig:IandI}, Social Intelligence also efficiently employs small data to generate big data, then refines the big data to produce deep intelligence from the perspective of the whole mining schedule. In addition, it helps to identify biases social and contextual factors that may affect the accuracy of data validation. By considering social aspects, Social Intelligence helps to improve the quality and credibility of the data validation process.

\textit{Ecological Intelligence}: It enables data validation from the framework perspective, which adopts an ecological level to solve complex tasks in intelligent systems. The development of autonomous transportation in open-pit mines significantly increases productivity. However, as data-driven models become more widespread, it is necessary to consider their wider personal, social, and ecological impacts beyond economic interests \cite{10399367}. When the diverse perspectives of these different agents are involved, the potential conflicting interests, demand a balanced solution harmonizing economic, social, and environmental factors. Ecological intelligence emphasizes the harmonization of data distribution across individual, group, and community levels, integrating natural, social, and networked environments, which delves into advanced ecosystem modeling, analysis, and management. Moreover, ecological intelligence adheres to principles of ethics, responsibility, and sustainability. In terms of data validation, ecological intelligence can provide a holistic method that ensures the integrity and validity of the validation process. By considering the broader ecological context, including interdependencies between heterogeneous data sources, the influence of external factors, and consistency with ethical and sustainability policies, ecological intelligence can contribute to robust and comprehensive data validation practices \cite{ERS}.

In summary, "3I" achieves the goal of data augmentation in terms of scale, quality, and structure, while the other "3I" accomplishes data validation in terms of safety, human recognition, and framework. The implementation of "6I" effectively improves the quality of data for autonomous transportation, establishing a solid foundation for autonomous transportation in open-pit mines. In addition, "6I" enables the realization of the ultimate "6S" goals of SE in open-pit mines: safety in the physical world, security in the cyber world, sustainability in the ecological world, sensitivity to the needs of individuals, service for all, and overall smartness. 

The integration of "6I" not only fulfills the "6S" goals but also signifies the evolution of data-driven models from feature-based components to scenario-based intelligent ecologies, which contributes to the intelligent evolution and sustainability of autonomous transportation systems. To effectively evaluate each of the "6S" objectives in autonomous transportation systems, specific metrics are designed for unique applications and functions. These metrics are used to assess and express the realization of each objective, forming another set of "6I" elements: Safety Index, Security Index, Sustainability Index, Sensitivity Index, Service Index and Intelligence Index, Service Index, and Intelligence Index.

\subsection{Calibration \& Certification}

C\&C (Calibration \& Certification) involves identifying suitable parameters for well-trained models to align with traffic participants in the real world in autonomous transportation and ensures the performance of these models through certification. Calibration manifests as minimizing the gap between model outputs and labeled instructions, and certification verifies the robustness and trustworthiness of a trained data-driven model. A credit evaluation system for autonomous transportation systems can be established through certification by a third-party organization \cite{SE1}. C\&C improves the robustness and trustworthiness of transportation systems in open-pit mines by aligning the virtual transportation system with the working conditions of a real scenario through parallel intelligence, which also provides proven performance guarantees.

\textcolor{black}{To improve autonomous transportation by the C\&C, several key steps should be implemented. First, learning potential relations from SFE for the entire scenario and coupling calibration to assist in determining optimal solutions for model parameters \cite{SE2}; Second, ensuring the internal dynamics of the data-driven model through parallel intelligence closely matches real-world operating conditions; Finally, certifying the autonomous driving system by a third-party organization and issuing a permanent certificate. }

The calibration process can be achieved by minimizing the total distance between the output of the trained model and the labeled commands. Certification aims to validate and confirm the technical performance of the calibration system, representing a formal recognition of the characteristics of the autonomous transportation system.

The certification must be issued by a reputable third-party organization. To fulfill the certification requirements, the following two criteria must be established:
\begin{itemize}
    \item Preservation: Certification should be easy to preserve, especially for systems designed for long-term operation. Digital certificates, especially cryptographic certificates in the form of non-fungible tokens in the blockchain, can be an ideal choice for long-term certification.

    \item \textcolor{black}{Uniqueness: Certification should be unique to clearly distinguish between different categories of the same system. This ensures that certified autonomous mining truck in open-pit mines is uniquely identified and distinguishable.}
\end{itemize}

When autonomous specialized equipment passes these established criteria, both certifications can be issued to confirm the robustness and trustworthiness of the autonomous transportation system. They can be manifested as non-transferable reputational endorsements or transfer of economic tokens, thus creating a trustworthy and robust data-driven system \cite{HE1}.

\subsection{Verification \& Validation}
Verification and Validation (V\&V) in autonomous transportation involves ensuring the robust and trustworthy performance of autonomous trucks. Verification focuses on assessing the compliance of autonomous transportation systems with the specifications through analyses, testing, and inspections, among other activities \cite{ge1}. It verifies that the trained model has been implemented correctly and meets the specified requirements. This includes methods such as code review, static analysis, and unit testing to verify the part components of the autonomous system. On the other hand, the validation evaluates the performance of the autonomous transportation system in real-world environments \cite{ge2}. This involves testing the functionality of the autonomous trucks in these challenging conditions to meet the specific requirements and expectations of open-pit mines. The validation process includes rigorous testing under actual working conditions and assessing the system's response to various challenges encountered in such scenarios. By validating the system's performance, any anomalies or gaps can be identified to improve overall robustness.

V\&V plays a pivotal role in identifying, mitigating, and resolving potential risks and issues in autonomous transportation systems. This process is instrumental in developing robust and trustworthy systems, which are essential for enhancing productivity, safety, and efficiency in open-pit mining scenarios.

\subsection{Organization of Autonomous Transportation Paradigm}
\textcolor{black}{To maintain the continuous optimization of the autonomous transportation paradigm, the organizational mechanisms have been carefully designed. As depicted in Fig. \ref{fig:framework}, including the federated contract and the federated incentive mechanism, which incorporate state-of-the-art algorithms into this paradigm. The federated contract records the rules of autonomous transportation in open-pit mines and automatically implements task-specific operations, such as agent state monitoring, execution permission checking, and automatic algorithm updating. }

For efficient and open-source algorithms, the passive Request-Permission (RP) mode and the active Automatic-Update (AU) mode are proposed. In RP mode, algorithm owners can request their algorithms to be included in the dynamic transportation paradigm, which will be further evaluated by V\&V to determine the permission status through a large number of mixed real and virtual experiments. In AU mode, the smart contract automatically monitors the status of papers and code in the auto-transport community, such as paper citations and code stars. When these metrics reach predefined thresholds, the system automatically initiates a request for algorithm inclusion. The joint incentive mechanism is designed to encourage developer participation. After the developed algorithms are incorporated into the framework of the autonomous transportation paradigm, ownership will be declared and revenue will be distributed to the owners based on the adoption rate of the algorithms.

\begin{figure}[b]
    \centering
	  \subfloat[Transportation Efficiency.]{
       \includegraphics[width=0.48\linewidth]{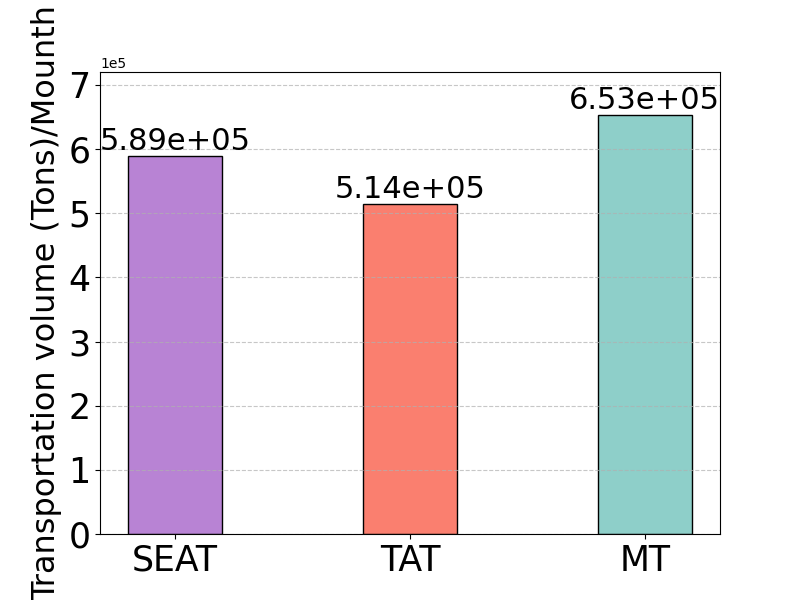}}
    \hfill
	  \subfloat[Transportation Safety.]{
        \includegraphics[width=0.48\linewidth]{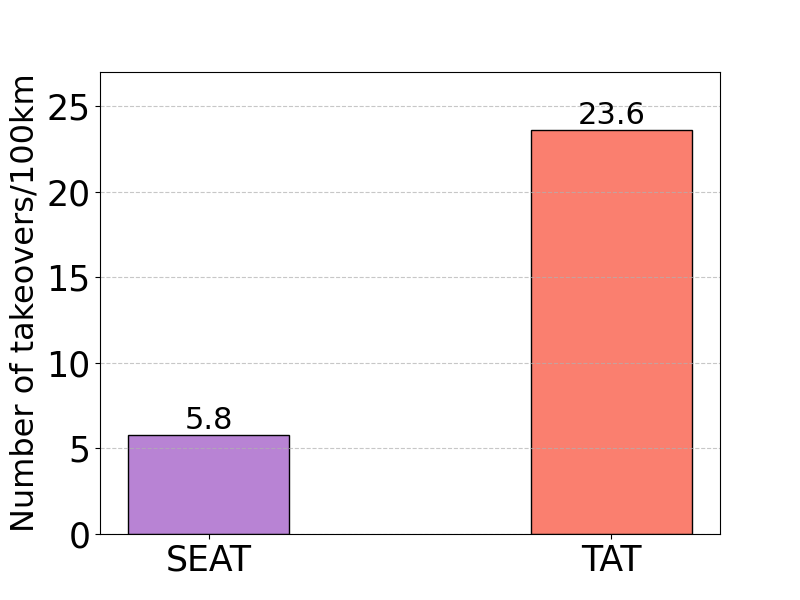}}
    \hfill
    \label{fig:firstmine}
    \caption{The comparison of three models in terms of safety and efficiency in Baorixile Open-pit Coal Mine}
\end{figure}

\section{Application and analization}
 One of the primary goals of SE is to enhance the robustness and trustworthiness of autonomous transportation systems in open-pit mines. This section demonstrates several two that implement SE in some famous open-pit mines, including the Baorixile open-pit coal mine and the Haerwusu open-pit coal mine.

In the experiment section, we provide a comparative analysis of the distinction in safety and efficiency between three different transportation systems in two open-pit mines, including \textbf{SEAT}, \textbf{TAT}, and \textbf{MT}. \textbf{SEAT} is the abbreviation for \textbf{SE}-based \textbf{A}utonomous \textbf{T}ransportation, which is proposed in this research. \textbf{TAT} is short for \textbf{T}raditional \textbf{A}utonomous \textbf{T}ransportation, which represents a traditional pipelined autonomous driving system. The full name of \textbf{MT} is \textbf{M}anual \textbf{T}ransportation, which stands for the metrics of transportation tasks performed by expert drivers.

% 在实验部分，我们着重对比了在这几个矿区中SE-integrated autonomous transportation(SEAT)与traditional autonomous transportation(TAT) 以及Manual transportation(MT)在安全性和效率方面的区别

% 实验部分的指标，一个是人工接管的次数，另外一个是运输效率。整体结果接管次数，SE>普通，效率是人工>SE>普通
 \subsection{Baorixile Open-pit Coal Mine}
% 宝日希勒露天煤矿位于呼伦呼伦贝尔草原东部，于1998年开始开采，其目前资源储量16.22亿吨，均为优质褐煤，煤层赋存条件十分稳定。该露天矿国家核定年生产能力为3500万吨，采用单斗—卡车工艺进行开采。由于地理原因，该露天矿冬天经常处于极寒条件中，最低温度达到了-50°C。

%极寒的工况也对自动运输提供了极大的挑战，如图所示，在安全性方面，SE-based的方法已经接管次数远远低于传统autonomous mining truck，其鲁棒性和可信性更强。在效率方面，相较于传统无人驾驶，SE的效率已经有了显著提升，但依然低于人类驾驶员。可以体现出，与传统方法相比，SE-based的方法已经有了限制提升，但其效率依然低于人工驾驶效率。

The Baorixile Open-Pit Coal Mine, is located in the eastern part of the Hulun Buir Grassland in the Intermolar province of China. The mine possesses an estimated coal reserve of 1.622 billion tons, predominantly comprising high-quality lignite, which has a production capacity of 35 million tons each year. Due to its geographical location, the mine often operates under extreme cold conditions during winter, with temperatures plummeting to as low as -50°C.

These adverse working conditions pose significant challenges to autonomous transportation systems. As indicated in Fig. \ref{fig:firstmine}, for the efficiency of autonomous transportation, although SEAT shows notable improvements over TAT, it still lags behind MT. In terms of safety, autonomous mining trucks with SEAT have substantially fewer takeovers compared to those with TAT, demonstrating enhanced robustness and reliability. This underscores that, while SEAT has achieved limited advancements over TAT, its efficiency remains inferior to that of MT.
\begin{figure}[t]
    \centering
	  \subfloat[Transportation Efficiency.]{
       \includegraphics[width=0.48\linewidth]{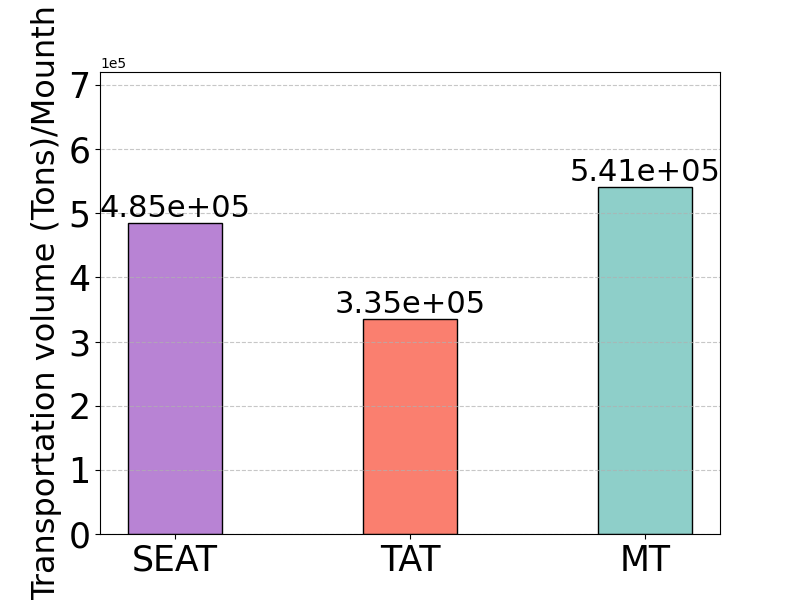}}
    \hfill
	  \subfloat[Transportation Safety.]{
        \includegraphics[width=0.48\linewidth]{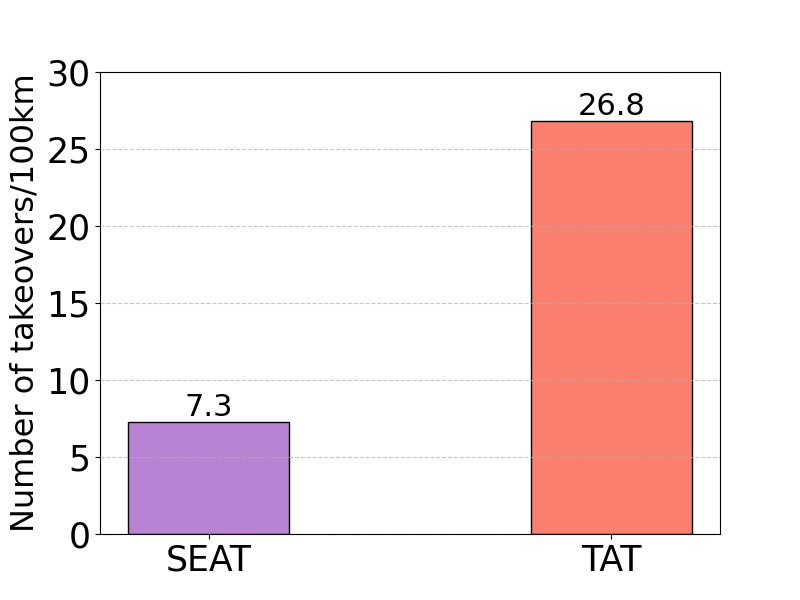}}
    \hfill
    \label{fig:secondmine}
    \caption{The comparison of three models in terms of safety and efficiency in Haerwusu Open-pit Coal Mine}
\end{figure}

 \subsection{Haerwusu Open-pit Coal Mine}
% Haerwusu Open-pit Coal Mine位于内蒙古自治区鄂尔多斯市，于2006年开始开采，其目前资源储量17.3亿吨，为世界上最大的露天矿山之一，其核定年生产能力为3500万吨。由于地理原因，该矿山是多风少雨的气候，这也导致其工作环境充满沙尘。这种工况也对自动运输提供了极大的挑战，如图所示，在安全性方面，SE-based的方法已经接管次数远远低于传统autonomous mining truck，其鲁棒性和可信性更强。在效率方面，相较于传统无人驾驶，SE的效率已经有了显著提升，但依然低于人类驾驶员。可以体现出，与传统方法相比，SE-based的方法已经有了限制提升，但其效率依然低于人工驾驶效率。

The Haerwusu Open-Pit Coal Mine, situated in the Ordos of the Inner Mongolia Province of China. With its current resource reserves estimated at 1.73 billion tons, it ranks among the largest open-pit mines in the world. The mine has an approved annual production capacity of 35 million tons. Geographically, the mine is located in a region characterized by high winds and scant rainfall, resulting in a work environment heavily laden with sand and dust. These conditions present substantial challenges for autonomous transportation systems.

As illustrated in Fig. \ref{fig:secondmine}, in terms of efficiency, while the SEAT has shown marked improvements over conventional autonomous driving, it still falls short of MT. From the safety perspective, the SEAT has significantly fewer interventions compared to TAT, indicating superior robustness and trustworthiness. This highlights that although SEAT has made limited advancements in comparison to TAT, its efficiency remains below that of MT.

It can be discerned from the experiment that, owing to the effective integration of SE with the autonomous transportation system, the SEAT demonstrates superior robustness and trustworthiness compared to the TAT. However, it remains susceptible to unforeseen scenarios, particularly in complex scenarios such as loading and dumping areas. This vulnerability necessitates remote takeover by experts, consequently impeding the efficiency of the transportation process.

\section{Conclusion}

In this research, an efficient transportation paradigm is proposed, which enhances the trustworthiness, robustness, and efficiency of the autonomous transportation system in open-pit mines by infusing scenario feature extractor, I\&I, C\&C, and V\&V within the SE framework. In the future, this paradigm will focus on increasingly harnessing the common sense capabilities of foundation models \cite{chat} to establish a human-centric operational system, which aims to offer intelligent services tailored to diverse scenarios encountered in open-pit mines.

% \section*{Acknowledgments}
% This should be a simple paragraph before the References to thank those individuals and institutions who have supported your work on this article.

% {\appendix[Proof of the Zonklar Equations]
% Use $\backslash${\tt{appendix}} if you have a single appendix:
% Do not use $\backslash${\tt{section}} anymore after $\backslash${\tt{appendix}}, only $\backslash${\tt{section*}}.
% If you have multiple appendixes use $\backslash${\tt{appendices}} then use $\backslash${\tt{section}} to start each appendix.
% You must declare a $\backslash${\tt{section}} before using any $\backslash${\tt{subsection}} or using $\backslash${\tt{label}} ($\backslash${\tt{appendices}} by itself
%  starts a section numbered zero.)}

\bibliographystyle{IEEEtran}
\bibliography{REf.bib}

\vfill

\end{document}